# Chapter 1

## Collision and Obstacle Avoidance for Industrial Autonomous Vehicles – Simulation and Experimentation Based on a Cooperative Approach

*J. Grosset, A.-J. Fougères, M. Djoko-Kouam, C. Couturier and J.-M. Bonnin*

### 1.1. Introduction

One of the challenges of Industry 4.0, is to determine and optimize the flow of data, products and materials in manufacturing companies. To realize these challenges, many solutions have been defined [1] such as the utilization of automated guided vehicles (AGVs). However, being guided is a handicap for these vehicles to fully meet the requirements of Industry 4.0 in terms of adaptability and flexibility: the autonomy of vehicles cannot be reduced to predetermined trajectories. Therefore, it is necessary to develop their autonomy. This will be possible by designing new generations of industrial autonomous vehicles (IAVs), in the form of intelligent and cooperative autonomous mobile robots.

In the field of road transport, research is very active to make the car autonomous. Many algorithms, solving problematic traffic situations similar to those that can occur in an industrial environment, can be transposed in the industrial field and therefore for IAVs. The technologies standardized in dedicated bodies (e.g., ETSI TC ITS), such as those concerning the exchange of messages between vehicles to increase their awareness or their ability to cooperate, can also be transposed to the industrial context.

---

J. Grosset
IMT Atlantique, IRISA, ECAM Rennes, Louis de Broglie, Campus de Ker Lann, Bruz, Rennes 35091, France

The deployment of intelligent autonomous vehicle fleets raises several challenges: acceptability by employees, vehicle location, traffic fluidity, vehicle perception of changing environments (dynamic), vehicle-infrastructure cooperation, or vehicles heterogeneity. In this context, developing the autonomy of IAVs requires a relevant working method. The identification of reusable or adaptable algorithms to the various problems raised by the increase in the autonomy of IAVs is not sufficient, it is also necessary to be able to model, to simulate, to test and to experiment with the proposed solutions. Simulation is essential since it allows both to adapt and to validate the algorithms, but also to design and to prepare the experiments.

To improve the autonomy of a fleet, we consider the approach relying on a collective intelligence to make the behaviours of vehicles adaptive. In this chapter, we will focus on a class of problems faced by IAVs related to collision and obstacle avoidance. Among these problems, we are particularly interested when two vehicles need to cross an intersection at the same time, known as a *deadlock situation*. But also, when obstacles are present in the aisles and need to be avoided by the vehicles safely.

This chapter is organized as follows: state of the art on algorithms and techniques to improve the autonomy of an IAV in Section 1.2. The improvement of a collision avoidance algorithm [2] in order to handle the problem of o is described in Section 1.3. In order to set up simulations and experiments, we propose an agent model which is presented in Section 1.4. The results related to the simulation are explained in Section 1.5. Section 1.6 describes the implementation of experiments with real robots. The results and a discussion of these experiments are initiated in Section 1.7. Finally, conclusions and future work are presented in Section 1.8.

## 1.2. State of the Art

In researches on Intelligent Transport System (ITS), autonomy of vehicles is well determined with 6 levels of autonomy [3]. However, no such scale exists in the industrial context, and too little research exists in this area [4, 5]. A few articles establishing a state of the art on the algorithms and techniques proposed to improve the control and relevance of the reactions of IAVs in the face of complex situations make it possible to verify the importance of this subject for Industry 4.0 [4, 6]. The study of these articles shows that more and more proposed solutions relate to decentralized control algorithms [6-12]. Among the problems to be solved to make IAVs more autonomous, we can note: task allocation [13-16], localization and vehicles positioning estimation [17-22], path planning [23, 24], motion planning [23, 25, 26] with particularly centralized collision avoidance [27, 28] and decentralized collision avoidance [24, 29, 30], deadlock avoidance [31-33], and vehicle resources management (battery for instance) [4, 34, 35].

The objective of our research is to improve the IAV autonomy integrated in a fleet based on collective intelligent strategies. The capacity to exchange information between the different IAVs of a fleet is necessary to improve this autonomy [36, 37]. Thus, the collision avoidance problem can be solved by the cooperation between IAVs [2, 24, 29, 30]. The study [2] proposed a cooperation strategy based on the exchange of messages to determine the priority to pass an intersection between IAVs. The solution requires the vehicle to know its own position, and to be able to communicate with the other vehicles. The collision avoidance algorithm presented in [2] allows IAVs to communicate and cooperate using different types of messages. The communication between IAVs is done with 3 different types of messages:

- *Hello_msg*: message to indicate its presence with its position;

- *Coop_msg*: message before an intersection area to determine priority;

- *Ack_msg*: message to confirm receipt of a *Coop_msg*.

The European Institute of Telecommunications Standards (ETSI) has published a standard for this kind of *Cooperative Awareness Message* (CAM) (ETSI EN 302 637-2 standard [38, 39]) and *Decentralized Environmental Notification Message* (DENM) (ETSI EN 302 637-3 standard [40]). These specifications and messages are approved and constitute building blocks for the safety of future ITS [41]. The purpose of CAM messages is similar to *Hello_msg* in [2]. Under, the assumption that each vehicle is able to localize itself (e.g., using GNSS), they allow to exchange positions, and thus activate cooperative awareness. Indeed, it allows the cooperative vehicles in the surrounding to be positioned in real time. This is based on a strong assumption: the vehicles must be able to locate themselves precisely. Localization is generally done with GPS, which is not very precise. Moreover, GPS does not work inside buildings, and so in our Industry 4.0 context, GPS is not the tool that IAVs will be able to locate themselves with. DENM messages are alert messages. They are issued at the time of an unexpected event in order to cooperate, warn and disseminate information in the geographical area concerned.

ETSI has also published a standard for *Cooperative Perception Messages* (CPM) (ETSI TR 103 562 standard [42]). They allow vehicles to broadcast information about objects perceived in their detection area by their sensors to other vehicles such as obstacles, pedestrians or other vehicles. Another way to cooperate is to inform other vehicles of these intentions. In this regard, the ITS WG1 is currently working on the definition of a *Maneuver Coordination Service* (MCS) and its associated *Maneuver Coordination Messages* (MCM) [43]. The outcome of this work item is planned for end of 2023. We expect MCM messages could be used or enhanced to schedule the access to crossroads.

CAM, DENM and CPM messages are important messages standardized. Therefore, we will propose a model of these messages adapted to the industrial context. Then, we will show their possible use to cooperate and avoid collisions for the IAVs with the example of the Bahnes et al. algorithm [2]. Furthermore, we will discuss the MCM messages and their possible use from an experimental perspective in the results and discussion section.

### 1.3. Algorithm Improvement

The collision avoidance algorithm of [2] makes it possible to deal with the priority of different vehicles when approaching an intersection. However, it does not deal with the problems of detection, communication and avoidance of fixed or moving obstacles (e.g., human operators).

In order to address the problem of fixed and mobile obstacles in warehouse aisles, we extended the algorithm of Bahnes et al. to handle the presence of fixed or moving obstacles, in our prior research [44]. In this context, we proposed two new types of messages for collaborative perception:

- *Obstacle_msg*: a message sent by an IAV agent to other IAV agents circulating in the warehouse to indicate the presence of a perceived obstacle;

- *Alert_msg*: a message sent by an IAV agent to other IAV agents circulating in the warehouse to indicate an unavoidable obstacle.

Then, we simulate the algorithm staying within the framework of the three scenarios proposed by [2]. These simulations rely on an agent-based model where IAVs are agentified [45, 46]. Indeed, agent-based simulation for IAVs is the most common in the same way as simulations based on discrete events or robotics software [47]. IAV agents have the ability to exchange messages and are equipped with radar. This allows them to detect vehicles in front of them. For instance, given an IAV agent $\alpha_i$, if another IAV agent $\alpha_j$ in front of it is stopped or is travelling at a slower speed, the IAV agent $\alpha_i$ can detect it with its radar and stop accordingly to avoid hitting it. To improve the collective autonomy of the IAVs it is essential that they have a good capacity for individual autonomy. The individual autonomy of the IAVs strengthens their collective autonomy.

The Bahnes algorithm proposed in our previous work [44] has been adapted with the standard ETSI messages used for ITS, in Fig. 1.1. These messages exchanged by the different IAVs remain consistent with Bahnes' algorithm and are adapted to the industrial context. The messages used that substitute the messages defined by Bahnes et al. are the CAM, MCM and ACK_MCM messages respectively to the *Hello_Msg*, *Coop_Msg* and *Ack_Msg* messages. Finally, the two types of message proposed for collaborative perception in our previous work [44], *Obstacle_msg* and *Alert_msg*, are replaced by CPM and DENM messages respectively.

The algorithm presented in Fig. 1.1 does not take into account the priority of passage in the case where 2 IAVs request passage at the same time at an intersection. This means that it does not propose a solution in case of a deadlock situation. We will discuss the different possibilities in the results and discussion section for our future work.

## 1.4. Modeling

In order to implement the Bahnes algorithm in simulation, we have modelled the different agents that will be implemented in the simulation. Fig. 1.2 shows the model agentified we have implemented in the simulation.

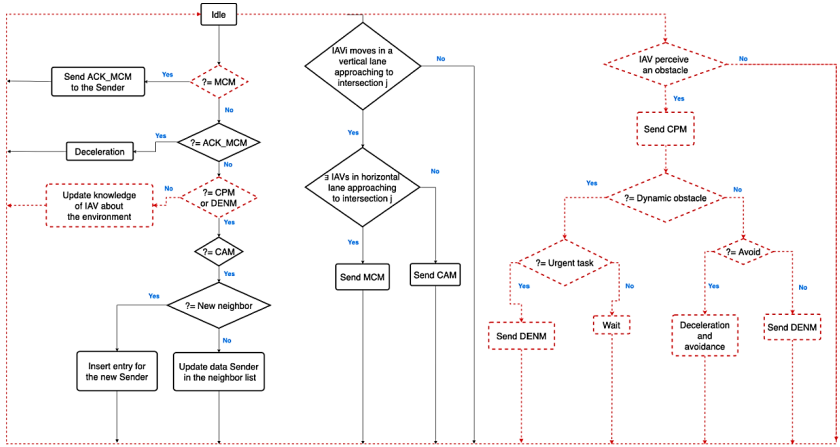

**Fig. 1.1.** Improvement of Bahnes's algorithm to treat the problem of collision and obstacles.

IAVs are equipped with radar to detect pedestrians, other IAVs (dynamic obstacles) and goods (static obstacles) present in their activity area. Then, to move in their environment and accomplish their mission, they have knowledge about their environment through their own perception of the environment and through the information received from other IAVs. Indeed, they exchange information with other IAVs or even with the infrastructure thanks to different types of messages: CAM, DENM, CPM, MCM and ACK_MCM. This allows IAVs to build up their own dynamic mapping of the environment. Thus they are cooperative, pro-active and autonomous to carry out their missions without colliding with static and dynamic obstacles.

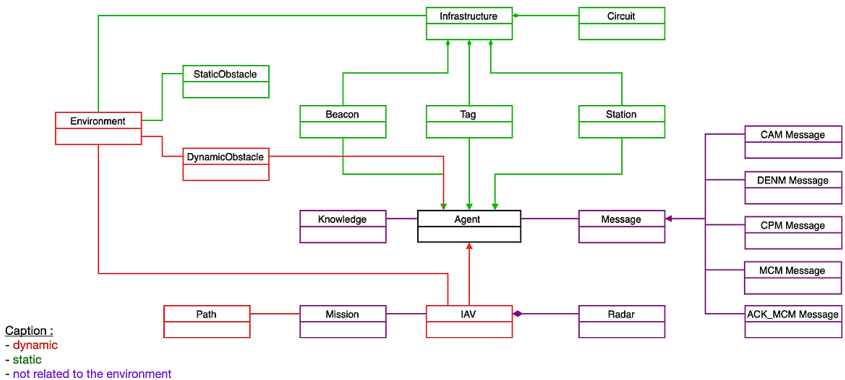

**Fig. 1.2.** Class diagram of the simulation.

The different messages presented in the Fig. 1.2 are used in our cooperation protocol for simulation and experiments. Their modelling and representation are inspired by the messages standardized in the TC ITS (Intelligent Transportation System) of the European Institute of Telecommunications Standards (ETSI) standard messages. That means, we transpose the messages of the Bahnes algorithm to equivalent ETSI messages we adapted to the industrial context which has different constraints than the road sector.

The purpose of *Hello_msg* proposed by Bahnes correspond to the *Cooperative Awareness Message* (CAM, ETSI EN 302 637-2 standard [38, 39]). This is a message sent by the vehicle to indicate its position in real time. We propose a model for the industrial context in Fig. 1.3 and its associated implementation as a message in ROS2 in Fig. 1.4.

| CAM Message | CAM | ItsPduHeader | |
|---|---|---|---|
| | | GenerationTime | |
| | | CAM Parameters | BasicContainer (CurrentPosition + StationType) |

**Fig. 1.3.** Representation of CAM Message for the industrial context.

| CAM Message | ItsPduHeader its_header | uint8 protocol_version | |
| | | uint8 message_id | CAM Message = 1<br>DENM Message = 2<br>CPM Message = 3<br>MCM Message = 4<br>ACK_MCM Message = 5 |
| | | uint32 station_id | |
| | uint16 generation_time | | |
| | StationType station_type | uint8 value | UNKNOWN = 0<br>PEDESTRIAN = 1<br>IAV = 2<br>BEACON = 3 |
| | float64[] current_position | | |

**Fig. 1.4.** Modeling of CAM Message in ROS2.

The alert message we proposed for the simulation of the augmented Bahnes algorithm will be implemented using the *Decentralized Event Notification Message* (DENM, ETSI EN 302 637-3 standard [40]). This can be of 3 different types: TRIGGER, UPDATE and TERMINATE (*message_type* in the modelling of the message in Fig. 1.6). TRIGGER corresponds to the first alert message, UPDATE to an alert message that allows to update information related to the first measurement, and finally the TERMINATE type that allows to inform others that the alert that has been issued is no longer active. This message contains the cause and sub-cause of the alert that is issued in the *SituationContainer* block of the message (see Fig. 1.5). Several alert message codes have been transposed from the standard for our experiments such as *Collision_RISK* with sub-causes associated with this code such as a longitudinal, lateral, intersection-related or vulnerable user collision risk (modelling of *SituationContainer* in Fig. 1.6).

| DENM Message | Decentralized Environmental Notification Message | DENM Parameters | ItsPduHeader |
|---|---|---|---|
| | | | Termination |
| | | | ManagementContainer (StationType + DetectionTime + Distance + ValidityDuration) |
| | | | SituationContainer (CauseCode + SubCauseCode + InformationQuality) |

**Fig. 1.5.** Representation of DENM Message for the industrial context.

| DENM Message | ItsPduHeader its_header | | uint8 protocol_version |
|---|---|---|---|
| | | | uint8 message_id |
| | | | CAM Message = 1<br>DENM Message = 2<br>CPM Message = 3<br>MCM Message = 4<br>ACK_MCM Message = 5 |
| | | | uint32 station_id |
| | | | uint8 message_type |
| | | | TRIGGER = 1<br>UPDATE = 2<br>TERMINATE = 3 |
| | ManagementContainer management_container | StationType station_type | uint8 value |
| | | | UNKNOWN = 0<br>PEDESTRIAN = 1<br>IAV = 2<br>BEACON = 3 |
| | | | uint64 detection_time |
| | | | float64 distance |
| | | | uint32 validity_duration (seconds) |
| | SituationContainer situation_container | CauseCode cause_code | uint8 value |
| | | | TRAFFIC_CONDITION = 1<br>ACCIDENT = 2<br>SLOW_VIA = 26<br>COLLISION_RISK = 97 |
| | | CauseCode sub_cause_code | UNAVAILABLE = 0<br>LONGITUDINAL_COLLISION_RISK = 1<br>CROSSING_COLLISION_RISK = 2<br>LATERAL_COLLISION_RISK = 3<br>INVOLVING_VULNERABLE_USER = 4 |
| | | | uint8 information_quality |
| | | | UNAVAILABLE = 0<br>LOWEST = 1<br>HIGHEST = 7 |

**Fig. 1.6.** Modeling of DENM Message in ROS2.

Bahnes' augmented algorithm allows vehicles to take obstacle detection into account. This augmentation has seen the arrival of a new message: *Obstacle_message* in our previous work [44]. ETSI has standardized a message for the ITS domain for the exchange of obstacle perception or other information called the *Cooperative Perception Message* (CPM, ETSI TR 103 562 [42]). Our CPM-inspired adaptation for the industrial context is shown in Fig. 1.7 and our ROS2 implementation is modeled in Fig. 1.8.

| CPM Message | Collective Perception Message | CPM Parameters | ItsPduHeader |
|---|---|---|---|
| | | | GenerationTime |
| | | | BasicContainer (CurrentPosition + StationType) |
| | | | SensorInformationContainer (type + confidence) |
| | | | PerceivedObjectContainer (objectID + distance + acceleration + yawAngle) |
| | | | NumberOfPerceivedObjects |

**Fig. 1.7.** Representation of CPM Message for the industrial context.

| CPM Message | ItsPduHeader its_header | uint8 protocol_version |
|---|---|---|
| | | uint8 message_id<br>CAM Message = 1<br>DENM Message = 2<br>CPM Message = 3<br>MCM Message = 4<br>ACK_MCM Message = 5 |
| | | uint32 station_id |
| | uint16 generation_time | |
| | StationType station_type | uint8 value<br>UNKNOWN = 0<br>PEDESTRIAN = 1<br>IAV = 2<br>BEACON = 3 |
| | float64[] current_position | |
| | SensorInformation sensor_information | uint8 type<br>UNKNOWN = 0<br>LIDAR = 1 |
| | | uint8 confidence<br>UNKNOWN = 0<br>LOW = 1<br>MEDIUM = 2<br>HIGH = 3 |
| | PerceiveObjectContainer perceive_object | uint8 objectID<br>UNKNOWN = 0<br>PEDESTRIAN = 1<br>IAV = 2<br>OBJECT = 3 |
| | | float64[] distance |
| | | float64[] acceleration |
| | | float64[] yaw_angle |

**Fig. 1.8.** Modeling of CPM Message in ROS2.

## 1.5. Simulation Results

Before moving on to real experiments, we simulate the algorithm proposed in Fig. 1.1 to verify that this cooperative solution is able to solve the collision and obstacle avoidance problems. This allowed us to validate our model proposed in the previous section, and to verify that other difficulties and problems do not appear before the real experimentation phase.

The simulation framework was taken from the Bahnes' algorithm. Therefore, our model presented in Fig. 1.2 was simulated with the different scenarios and the traffic plan proposed by Bahnes et al. [2] presented in Figs. 1.9, 1.10. Indeed, this choice is justified by our conviction to use it as a benchmark plan to compare results.

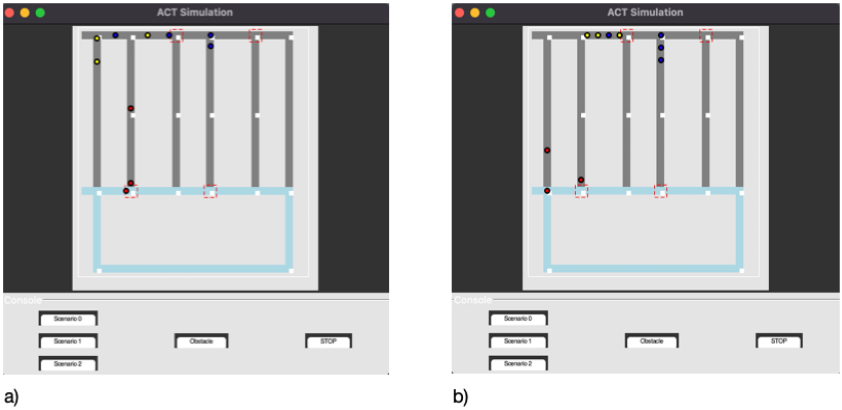

**Fig. 1.9.** Simulation of radar use: a) at the top of the picture: one blue and three yellow IAVs arrive near the intersection, b) while waiting for the yellow IAV to pass the intersection, the radar of the blue IAV and the two others yellow IAV allow them to stop and keep their distance to avoid colliding.

Ten IAVs start from the horizontal central aisle on the left. The red IAVs do the first circuit corresponding to the first loop at the top left, the blue ones follow them but will go down the $4^{th}$ aisle ($3^{rd}$ loop). Finally, the yellow ones do the big loop going clockwise just like the red and yellow IAVs. The upper part of this traffic plan illustrates the possibility of checking the correct management of the intersection. In addition, our simulation allows us to place obstacles at random positions in the lanes. This makes it possible to check the correct reaction of the IAVs in relation to their perception of the obstacles and the logs of their message exchanges. It involves different intersections, where vehicles can arrive from different sides like in a warehouse (4 intersections are shown in Fig. 1.10). Thus, it provides the different characteristics of an industrial environment and allows us to realize simulated experimental tests in line with realistic scenarios.

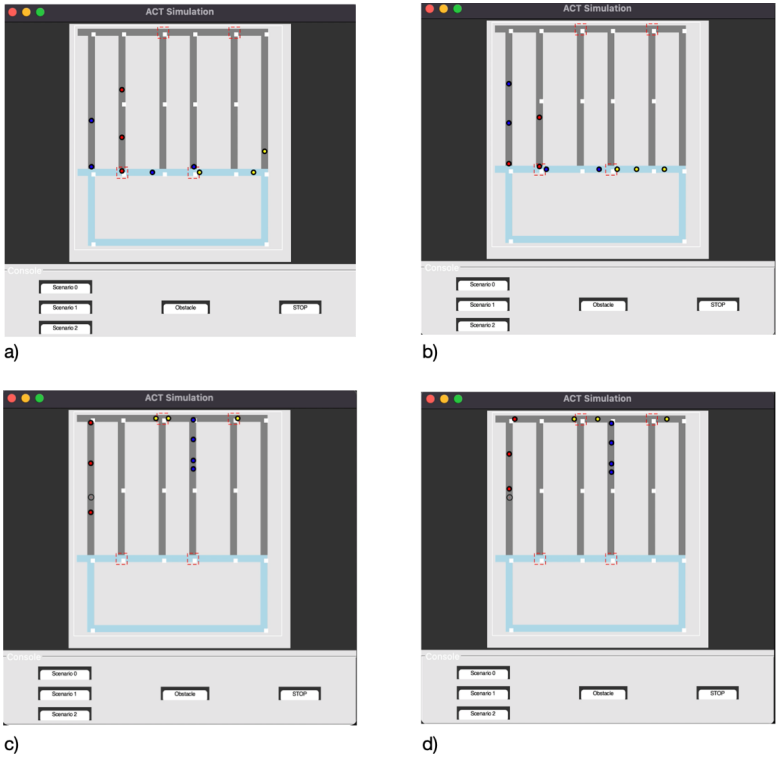

**Fig. 1.10.** Simulation of the scenarios: a) in the center of the picture: a blue and yellow IAV arrive at an intersection, b) the yellow IAV passed the intersection after communicating with other IAVs, c) on the left side of the picture: a red IAV perceives a fixed obstacle in front of him, d) a red IAV avoided the obstacle.

We notice in the simulation that the avoidance is well respected, and the obstacles are perceived by the IAVs. Therefore, the simulation validates the extended Bahnes's algorithm with collision avoidance and fixed or dynamic obstacle detection processing.

## 1.6. Experimentations

### 1.6.1. Context

After having carried out simulations, the objective is to move on to the stage of real experiments to test cooperation algorithms. As a first step, we want to verify that the augmented Bahnes algorithm presented in Fig. 1.1 does indeed allow industrial vehicles to avoid collisions with obstacles and to manage decision making at intersections. The main objective is to validate our hypothesis that collective intelligence strategies between vehicles will increase their individual and collective autonomy to perform their tasks efficiently.

In the Industry 4.0 context, we use Turtlebot3 'burger' robot as our representation of autonomous industrial vehicles. These robots are equipped with different components as described in Fig. 1.11 and use a *Raspberry Pi* and the *Robot Operating System* (ROS). Indeed, ROS is an open-source framework for the development of robotics applications and is the tool favored by researchers and even industrialists today. For our research we use ROS2 because it provides real-time control systems and large-scale distributed architectures [48]. Compared to ROS, there is no master entity, and ROS2 utilized Data Distribution Service (DDS). Thus, as we are working on distributed robotic systems to evaluate our cooperation algorithms, the choice focused on ROS2 naturally for our robot fleets. To facilitate real experiments, we first used the Gazebo simulation environment which is easily integrated with ROS2 and where Turtlebot3 can be simulated. This part of the simulation in the Gazebo environment makes it possible to check the correct operation of message exchanges and the cooperation algorithm and thus avoid collisions in experiments with physical robots. In this simulation environment, we first created a world representing an intersection where we deployed 4 Turtlebot3 'burger' equipped with their LIDAR (see Fig. 1.11).

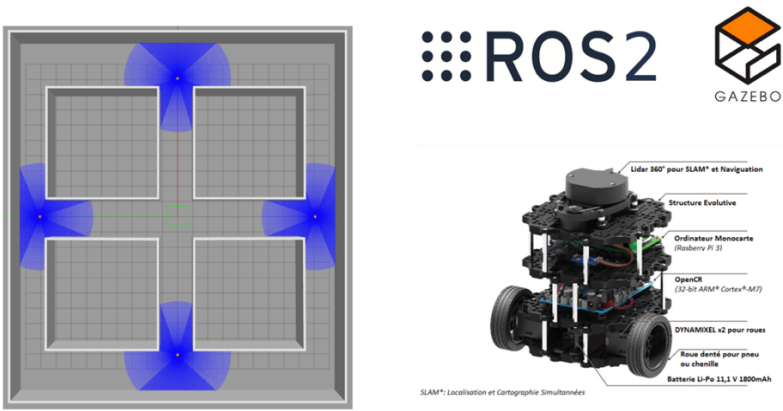

**Fig. 1.11.** Simulation environment for experimentations.

### 1.6.2. Collective Cooperation Strategy

Before simulating the Bahnes algorithm or other cooperative algorithms with standardized messages with the ROS2 framework and Gazebo, we worked on the task assignment for autonomous vehicles to implement robot control and movement utility related to the notion of tasks/missions.

We have assumed that the tasks are known and therefore configured in a file with the destination of each robot for each task. The destinations where the robots must perform tasks are represented by *PositionAction* which is (x, y) coordinates relative to the 2D world simulated in Gazebo. Our robotic architecture is similar to that used by Choirbot [49], a ROS2 toolbox for cooperative robotics.

That is, we have a layer to guide and a layer to plan the destinations of the robots. Indeed, each robot is associated to a *planner_client* node which sends the *PositionGoal* related to the destination of the task associated to the robot. Then, the *planner_server* node which allows us to standardize this destination point which it publishes to the goal topic associated to the robot. Afterwards, the guidance layer will subscribe to this topic to control the robot until it reaches its mission position. Once the robot has reached its destination, the odometry server sends feedback to the *planner_client*, which can reply the next *PositionAction* to the *planner_server*.

The communication between the different nodes for the control of robot 1 for example is shown in Fig. 1.12.

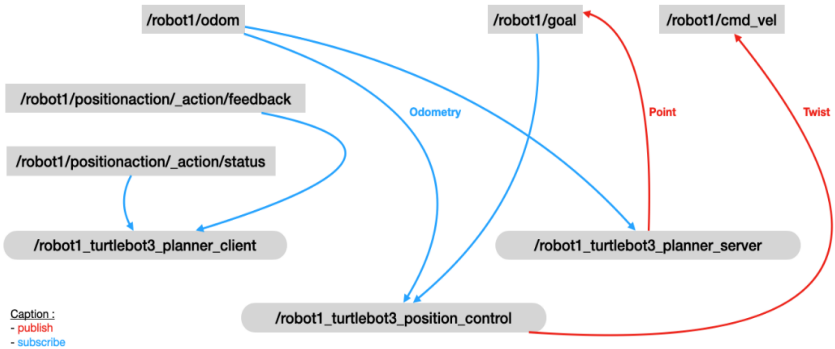

**Fig. 1.12.** Nodes and topics representation for robot1 for path planning.

Once the tasks have been assigned and the robots have been controlled, the robots must be able to communicate. This will allow the exchange of information on their positions, their vision of the environment, or their trajectory intention in order not to collide with each other or with obstacles.

### 1.6.3. Communication

In order to implement algorithms for collective cooperation between autonomous vehicles, we have implemented the different standardized messages presented in Section 1.4. The implementation of those messages in ROS2 will allow robots to exchange these types of messages through topics.

The *turtlebot3_position_control* node of a robot allows it to control its speed and orientation towards its destination point defined by the */goal* topic as explained in Subsection 1.6.2. This node will also allow to exchange messages related to its observations of the environment thanks to its LIDAR. We have defined an observation distance and a safety distance. When the LIDAR detects something corresponding to the observation distance or less it publishes a CPM message in the topic corresponding to the robot with the related information. Similarly, if it detects something at a distance less than its safety distance, it sends an alert message, i.e., a DENM message with the information associated with the DENM topic. This process of subscribing and publishing messages similar to all robots is modelled in Fig. 1.13 using robot 1 as an example.

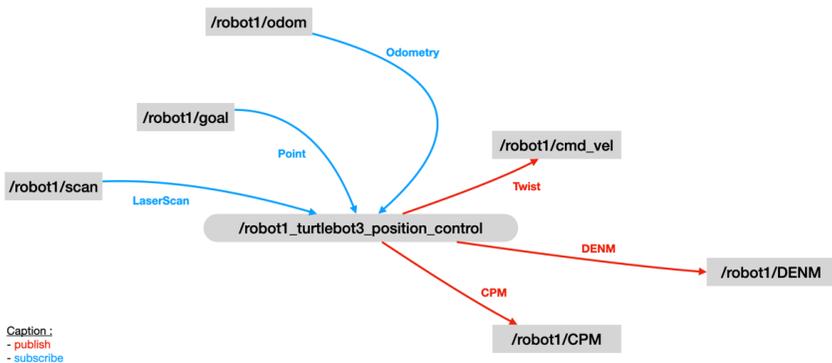

**Fig. 1.13.** Process of publishing/subscribing topics for position_control node of each robot.

CAM messages are published in CAM topics similar to CPM and DENM messages. They are standardized using information from the robot's odometry sensor. This information published by one robot is retrieved by all other robots in the manner of a broadcast exchange. The architecture has been implemented so that each robot has an *exchange_messages* node which allows it to subscribe to all the message topics of the other robots, i.e., CAM, CPM and DENM messages in our current experiment. These exchanges are modelled in Fig. 1.14 for 2 robots, but the process is similar regardless of the number of robots.

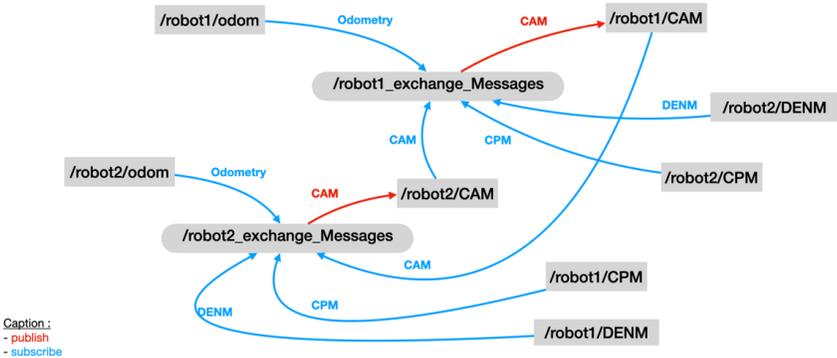

**Fig. 1.14.** Nodes and topics representation for exchanging messages.

## 1.7. Results and Discussion

Before actually experimenting with the strategic cooperation of the robots to avoid real collisions, we tested a simple scenario in a gazebo simulation. Each robot was given the task of crossing the intersection shown in Fig. 1.11, and they were asked to go to the *PositionGoal* in front of them.

We will take Fig. 1.15 as an example scenario. The robots broadcast their positions using CAM messages while moving towards the intersection. The LIDAR of robot2 detects an obstacle and sends a CPM message to share its information. But as we have not implemented obstacle avoidance control, it will stop at the safety distance of the LIDAR and send a DENM message with a risk of longitudinal collision. Similarly, robot3 and robot4 did not encounter any problems on their paths but when crossing the intersection, their LIDAR detects at the safe distance the other robot with an angle of less than 45°. They will immediately send a DENM message of risk of longitudinal collision as well.

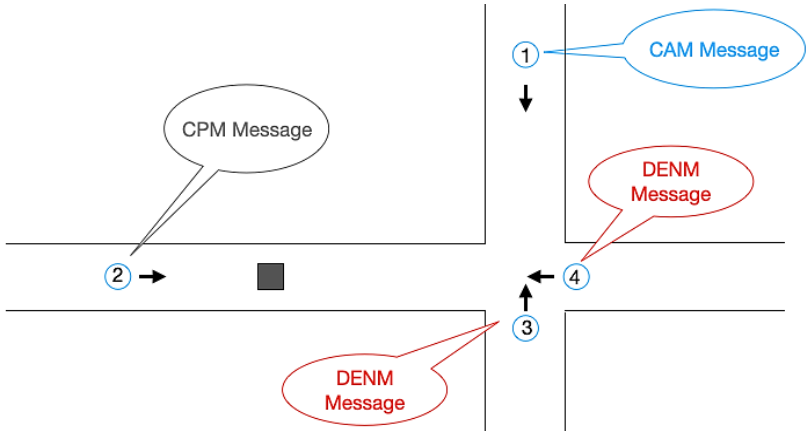

**Fig. 1.15.** Scenarios of message exchanging in experiments.

It can be noted that unlike robot2, robot3 and robot4 did not send a CPM message before, because at the moment of the intersection they were already too close. Robot3 and robot4 stop to avoid collision. Our goal was to find a cooperative solution for VIAs to avoid collisions and obstacles, but also to avoid unnecessary braking and stopping to optimise energy and speed. As a first step, these results show that our cooperative strategy allows the robots not to collide with each other or with obstacles.

Our results of communication between the different robots are therefore validated in the Gazebo simulation environment. Nevertheless, in order to be able to find a cooperation to cross the intersection it would be necessary to implement the cooperative messages presented by the Bahnes algorithm. For this purpose, a service not yet implemented by ITS WG1 and ETSI is discussed: *Maneuver Coordination Service* (MCS) and its associated *Maneuver Coordination Messages* (MCM) [43]. We then propose in the same way a representation for the industrial context in Fig. 1.16 and a modelling of the message for ROS2 in Fig. 1.17. This message would share the information that one wishes to cross an intersection by indicating *ManeuverContainer* information, that is *id* of the intersection, as well as the direction one would take.

| MCM Message | ManeuverCoordination | ItsPduHeader | |
|---|---|---|---|
| | | | GenerationTime |
| | | MCM Parameters | BasicContainer (CurrentPosition + StationType) |
| | | | ManeuverContainer (ReferenceIntersection + Direction) |

**Fig. 1.16.** Representation of MCM Message for the industrial context.

| MCM Message | ItsPduHeader its_header | uint8 protocol_version |
|---|---|---|
| | | uint8 message_id<br><br>CAM Message = 1<br>DENM Message = 2<br>CPM Message = 3<br>MCM Message = 4<br>ACK_MCM Message = 5 |
| | | uint32 station_id |
| | uint16 generation_time | |
| | StationType station_type | uint8 value<br><br>UNKNOWN = 0<br>PEDESTRIAN = 1<br>IAV = 2<br>BEACON = 3 |
| | float64[] current_position | |
| | ManeuverContainer maneuver | uint8 id_intersection |
| | | uint8 direction<br><br>STRAIGHT = 0<br>LEFT = 1<br>RIGHT = 2 |

**Fig. 1.17.** Modeling of MCM Message in ROS2.

Thus, in a future work, we would assume that the vehicles know the positions of the different intersections, or are able to locate them, and send an MCM message to the other robots indicating their planned trajectory in the intersection. The robots concerned by the request will be able to respond to an *ACK_Message* indicating their agreement or disagreement with the request. The industrial representation of this message and the ROS2 modelling we propose are detailed in Figs. 1.18, 1.19.

This discussion around MCM messages to enable vehicle cooperation when approaching an intersection raises an issue: if several vehicles request to cross the same intersection at the same time, or if one vehicle indicates that it does not agree to the request of another vehicle, a deadlock situation arises. There are 3 possibilities to manage this concern. The first is the idea of strong cooperation, i.e., all vehicles always agree. The second is to set up a supervisor who can arbitrate the situation and therefore choose which vehicle should pass. This case takes us out of a distributed architecture. Finally, the last solution is to add an algorithmic layer that is known to all the robots, and which therefore serves as a decisive judgment. All robots should take the same decision in a given situation. For example, if one robot disagrees with the crossing of another, it is then the level of priority and urgency of the task between the two robots that will decide who will be the first to cross. This solution would make it possible to always remain in the idea of strong collaboration, cooperation between the different agents in the situation.

| ACK_MCM Message | AckManeuverCoordination | ItsPduHeader | |
|---|---|---|---|
| | | GenerationTime | |
| | | ACK_MCM Parameters | BasicContainer (CurrentPosition + StationType) |
| | | | DestinaterContainer (StationType + StationID) |
| | | | ManeuverContainer (ReferenceIntersection + Direction) |
| | | | AckResponse |

**Fig. 1.18.** Representation of ACK_MCM Message for the industrial context.

| ACK_MCM Message | ItsPduHeader its_header | uint8 protocol_version |
| --- | --- | --- |
| | | uint8 message_id |
| | | CAM Message = 1<br>DENM Message = 2<br>CPM Message = 3<br>MCM Message = 4<br>ACK_MCM Message = 5 |
| | | uint32 station_id |
| | uint16 generation_time | |
| | StationType station_type | uint8 value |
| | | UNKNOWN = 0<br>PEDESTRIAN = 1<br>IAV = 2<br>BEACON = 3 |
| | float64[] current_position | |
| | StationType station_type_destinater | uint8 value |
| | | UNKNOWN = 0<br>PEDESTRIAN = 1<br>IAV = 2<br>BEACON = 3 |
| | uint32 station_id_destinater | |
| | ManeuverContainer maneuver | uint8 id_intersection |
| | | uint8 direction |
| | | STRAIGHT = 0<br>LEFT = 1<br>RIGHT = 2 |
| | bool ack_mcm_response | |

**Fig. 1.19.** Modeling of ACK_MCM Message in ROS2.

## 1.8. Conclusions and Future Work

In an Industry 4.0 context, many actors cross paths in different areas of a warehouse or a factory: vehicles, operators, obstacles (objects that fall or left in the aisles may appear).

A specific state of the art on the used a message-based communication protocol between vehicles to prioritise the passage through an intersection allowed us to identify the Bahne's algorithm [2], well representative of the cooperative strategies developed in the field. As a first step, we carried out an algorithmic work to extend this algorithm in order to have the possibility to manage the detection of fixed and mobile obstacles (Fig. 1.1). Then, we proposed an agent model as well as a model adapted for the Industry 4.0 context of the ETSI standard messages for ITS. This proposed cooperation protocol is implemented for the exchange of information on location, perception of the environment, and notification of dangerous events.

We validated the extended algorithm by a simulation approach with a traffic plan presented in the literature. Finally, we emulated these different exchanges of awareness and perception messages in a virtual world of Turtlebot3 'burger' robots with ROS2 and Gazebo.

As an extension of this work and in order to perform real experiments, we discussed these results as well as a cooperation message named *Maneuver Coordination Messages* (MCM). This implementation perspective would allow us to validate the Bahnes algorithm augmented with the exchange of cooperation messages to describe one's intention to cross an intersection. In a second step, we aim at implementing an algorithmic layer for decision making in case of deadlock situation as discussed in Section 1.7.

Finally, we also plan to involve the intersection proximity infrastructure in the exchange of communication through camera information in order to help the VIAs to cooperate in crossing the intersection by avoiding unnecessary braking and stopping. This would optimize the energy and efficiency with which the VIAs carry out their missions.

## Acknowledgements

The authors would like to thank the region Bretagne-France for funding the VIASIC project as part of the ARED-2021-2024 call for projects concerning the strategic innovation area: Economics of industry for intelligent production.